# Enhanced Multi-Robot SLAM System with Cross-Validation Matching and Exponential Threshold Keyframe Selection

Ang He, Xi-mei Wu, Xiao-bin Guo, Li-bin Liu

**Abstract**

The evolving field of mobile robotics has indeed increased the demand for simultaneous localization and mapping (SLAM) systems. To augment the localization accuracy and mapping efficacy of SLAM, we refined the core module of the SLAM system. Within the feature matching phase, we introduced cross-validation matching to filter out mismatches. In the keyframe selection strategy, an exponential threshold function is constructed to quantify the keyframe selection process. Compared with a single robot, the multi-robot collaborative SLAM (CSLAM) system substantially improves task execution efficiency and robustness. By employing a centralized structure, we formulate a multi-robot SLAM system and design a coarse-to-fine matching approach for multi-map point cloud registration. Our system, built upon ORB-SLAM3, underwent extensive evaluation utilizing the TUM RGB-D, EuRoC MAV, and TUM_VI datasets. The experimental results demonstrate a significant improvement in the positioning accuracy and mapping quality of our enhanced algorithm compared to those of ORB-SLAM3, with a 12.90% reduction in the absolute trajectory error.

**Keywords**: SLAM; feature matching; keyframe; CSLAM; coarse-to-fine matching

## 1. Introduction

Rapid advancements in simultaneous localization and mapping (SLAM) technology have greatly accelerated progress in autonomous driving, augmented reality, and 3D re-construction, which enable mapping of unknown scenes and determining the location of sensors in real time. Camera-based real-time map building systems, such as visual odometry (VO) and visual simultaneous localization and mapping (VSLAM), are recognized as viable alternatives to LiDAR-based systems due to their ability to map unknown scenes and determine sensor locations in real time. These camera-based systems offer richer scene information and are lightweight, cost-effective, energy-efficient, and compact [1], making them essential technologies for current smart mobile devices, including virtual reality (VR) and augmented reality (AR) devices.

In the past two decades, both the theory and application of VSLAM have rapidly advanced, resulting in the development of numerous processing algorithms. Compared with other algorithms, which include the indirect method and extended Kalman filter (EKF)-based Mono-SLAM [2], semidirect visual odometry (SVO) [3] integrates both feature-based and direct methods, the visual-inertial navigation system (VINS) [4] employs Lucas-Kanade for feature tracking and DBoW2 for loop detection and integrates 4-DOF bitmap optimization and map merging; additionally, direct sparse odometry (DSO) [5] combines direct methods with sparse reconstruction. ORB-SLAM3 [6] exhibits significant advantages in terms of accuracy and robustness. It provides a SLAM system solution for monocular, stereo, and RGB-D cameras with visual and visual-inertial multiple-map capabilities, offering an abstract camera representation mode compatible with various types of camera sensors.

Feature matching is a fundamental visual processing technology that serves as the cornerstone of front-end image processing in SLAM systems. However, existing SLAM design schemes typically employ single-step verification during feature matching, which can lead to potential mismatches. The keyframe selection process reduces data redundancy, effectively easing the computational burden while ensuring the accuracy and reliability of the SLAM system. Despite the importance of keyframe selection, quantitative evaluations of existing SLAM systems are lacking. While sparse point cloud maps meet basic SLAM requirements, globally consistent dense maps offer more detailed geometric information, intuitive visualizations, and adaptability to complex scenes. Compared to single-robot SLAM methods, the multi-robot collaborative SLAM [7] [8] scheme is more efficient. However, current multi-robot SLAM schemes often overlook prior information from the initial state, resulting in high computational complexity. For the above issue, we propose a novel feature matching and keyframe selection approach that incorporates cross-validation into the feature matching process and constructs an exponential threshold function for keyframe determination, aiming to improve positioning accuracy and high-precision map reconstruction. The main contributions are as follows:

(1) A novel cross-validation approach is introduced for SLAM image feature matching to enhance matching accuracy by filtering mismatching points.

(2) By integrating keyframe selection strategies from structure recovery from motion and data association, an exponential threshold function is used for key frame selection. Quantizing the process improves system accuracy and mapping effectiveness by ac-counting for various influencing factors.

(3) Leveraging prior robot state information, a coarse-to-fine multi-map matching scheme for globally consistent dense map construction is developed.

**2. Related works**

In this section, we discuss related work on feature matching, keyframe selection, and multi-map construction.

2.1. Feature matching

Before feature matching, distinct and identifiable salient point structures must be extracted from the images. These structures, often corners, intersection points, or centroids of enclosed regions, are typically obtained through response function construction [9] [10] or sparse sampling from feature lines or contours [11]. Commonly used extraction methods include Harris [12], FAST [13], FAST-ER [14], AGAST [15], FAST [13], SIFT [16] [17], SURF [18], and ORB [19].

Once two sets of matchable feature points are extracted, image matching involves pairing these sets. Feature matching methods can be broadly categorized into direct and indirect approaches. Direct matching estimates correspondence points directly, while indirect matching establishes initial correspondences based on feature description similarity and subsequently eliminates mismatches through geometric constraints. Additionally, the emergence of deep learning technology, particularly deep convolutional networks, has sparked significant interest in feature matching methods that leverage deep feature layers. However, due to operational speed limitations, widespread adoption of these methods has been constrained.

Direct matching, or pure point set matching, establishes matching relationships between two point sets through constraints and objective function optimization. Correspondences matrix estimation methods aim to register point sets by minimizing the objective function and incorporating space distance and assignment matrix considerations while considering the smoothness and complexity of the transformation function. For scenes with nonrigid transformations, radial basis function [20] interpolation is commonly used. However, these methods may suffer from performance degradation with outliers, data degradation, and high solution space complexity. In contrast, graph model-based feature matching treats feature points as graph vertices [21] [22]. The key steps include graph construction, edge definition, and matrix construction. Optimization methods, which treat matching as a quadratic assignment problem (QAP) [22] [23], aim to solve it using gradient and spectral methods. Despite its theoretical value, optimizing graph matching for speed and robustness remains a key research direction due to its computational complexity and noise sensitivity.

The indirect feature matching strategy assigns feature point descriptors to establish preliminary correspondences, with common methods including floating-point (e.g., SIFT) and binary (e.g., BRIEF) descriptors. SIFT constructs robust high-dimensional vectors based on local gradient directions, while BRIEF achieves speed through pixel comparisons. Combining methods such as ORB with FAST and Harris response using BRIEF can quickly establish initial matches. However, noise, outliers, and occlusions often lead to mismatches, with existing methods having a mismatching rate of up to 50% [24]. Stricter thresholds can improve correct match ratios but may sacrifice valid matches. False matches are eliminated using methods such as RANSAC [25], MLESAC [26], and PROSAC [27]. or nonparametric interpolation. Despite advances in relaxation methods such as LPM [28] and GMS [29], balancing accuracy, handling outliers, and adapting to complex transformations remain challenges. Quickly establishing initial matches and designing accurate mismatch elimination strategies are key research problems in this area.

ORB-SLAM3 [6] employs an indirect feature matching method, leveraging the ORB algorithm for feature extraction and implementing various acceleration techniques for real-time performance. These methods can be divided into three categories. (1) Feature matching acceleration based on grid registration information facilitates 3D-2D matches by utilizing feature point grid registration information to find matching points near the 3D point projection position. (2) Feature matching acceleration based on bag-of-words vectors enables 2D-2D matches between key frames and ordinary frames by traversing only feature points belonging to the same node to find matches. (3) The acceleration of feature matching based on stripe search is particularly applicable to stereo cameras under the assumption of horizontal epipolar lines. Here, feature points from stereo images are aligned, and sliding window and subpixel matching are performed for refined correspondences. Through these acceleration techniques, ORB-SLAM3 efficiently identifies matching points. Building upon this, in feature matching acceleration based on grid registration information, we utilize the known transformation matrix between two frames to perform the inverse transformation for cross-validation. This approach effectively reduces the false matching rate, achieving a commendable balance between matching speed and accuracy.

2.2. Selection of the keyframe

Keyframe selection, explored across diverse fields, includes methods such as deep neural networks [30], but their reliance on training data limits their use in computer vision. Other methods, such as cluster-based [31] or shot-based [32] approaches in video summarization, require all frame data initially, hindering real-time application in computer vision algorithms. Many studies have focused on keyframe selection in visual odometry (VO) and visual SLAM, generally categorized into structure from motion (SFM) recovery and data association.

Methods for structure from motion (SFM) keyframe selection fall into five categories: same time or interval, overlapping images, disparity, information entropy, and image content. Examples include PTAM [33], SVO [3], LSD-SLAM [34] and RGB-D SLAM [35] for temporal or spatial thresholds. RANSAC in FAST-SLAM [36] introduces a method based on average image pixel displacement. OKVIS [37] selects keyframes based on matching point area ratios. VINS-mono [38] considers average disparity and tracking quality. Kerl *et al*. [39] used differential entropy. Image content methods establish a feature clustering space and select keyframes based on feature distance, although accuracy is challenging [40].

Fanfani *et al.* [41] proposed a data association-based keyframe selection method emphasizing high temporal difference feature ratios, enhancing robustness compared to threshold-based approaches. Engel *et al*. [42] presented a keyframe selection method suitable for low-light conditions by rapidly selecting optimal keyframes by marginalizing redundant keyframes and integrating sudden viewpoint changes. Chen *et al*. [43] suggested keyframe selection based on view change magnitude and rate, ensuring good localization accuracy despite constructing a sparse map.

ORB-SLAM3 [6] employs a key-frame selection strategy similar to ORB-SLAM2, frequently inserting keyframes and removing redundant ones. Keyframes are added to the local mapping thread if they meet specific criteria: maintaining a certain frame interval with the preceding keyframe, exhibiting sufficient spatial distance from the previous keyframe, having a decreased number of coview map points with the previous keyframe, having a sufficient number of successfully tracked feature points in the current frame, and ensuring that the local mapping thread is idle. Despite offering higher accuracy and robustness, frequent keyframe insertion consumes significant system resources, and the lack of overall quantitative evaluation affects positioning and mapping effectiveness.

2.3. Multi-robot map building

Compared with a single robot, a system composed of multiple robots has significant advantages, such as strong adaptability to the environment, strong carrying capacity, strong robustness, and high operating efficiency. Multi-robot CSLAM methods can be divided into centralized, distributed, and hybrid methods. Researchers have focused mainly on the communication between multiple robots, the accuracy of cooperative localization and the consistency of map fusion. Related work has focused mainly on solving the front-end (such as inter-robot closed-loop identification) and back-end (such as distributed pose graph optimization) problems of multi-robot CSLAMs. Inter-robot loop closure is pivotal for aligning the trajectories of robots in a shared reference frame, enhancing their trajectory estimation accuracy. In centralized approaches, a common approach to identifying closed loops involves visual place recognition techniques based on image or keypoint descriptors [44] [45]. Pose graph optimization (PGO) for multirobot systems is centrally managed by a central station using a centralized back-end approach [46] [47] [48]. Among the distributed methods, distributed pose graph optimization [8] [49] [50] and distributed factor graph solvers [51] [52] are employed. The hybrid structure combines the centralized structure and the distributed structure. The hybrid structure contains a unified global central management module, and the communication mode between each member robot adopts a distributed structure. Frontend and backend algorithms have also been demonstrated in comprehensive CSLAM systems such as [7] [8] [53] [54]. However, these methods do not make full use of the prior information in the initial stage to improve the robustness of the system and enhance the overall amount of calculation.

**3. The proposed method**

3.1. System Overview

Based on the current advanced ORB-SLAM3, the proposed system also implements three primary threads: loop detection, local mapping, and tracking. The overall operating framework is shown in Fig. 1. By employing cross-validation to filter out erroneous matching points, the precision of feature matching is enhanced. Additionally, the use of an exponential threshold function in the keyframe selection strategy quantifies the keyframe selection process. Concurrently, we introduce a dense mapping thread to facilitate the generation of the local point cloud map. To address the multi-robot protocol SLAM problem, we adopt a centralized structure where the master control unit communicates with each robot and performs the following actions: (i) collect the point cloud information generated by a single robot and process it through coarse-to-fine matching, uniform sampling, and Gaussian filtering; (ii) collect the local trajectory data and perform local trajectory fusion; and (iii) perform motion planning operations to control the motion of each robot.

In the rest of this section, we introduce our improvements: the integration of cross-validation for feature point matching, the implementation of an exponential thresh-old function within the frame processing stage to determine key frames, and the incorporation of a coarse-to-fine matching approach within the map data processing stage.

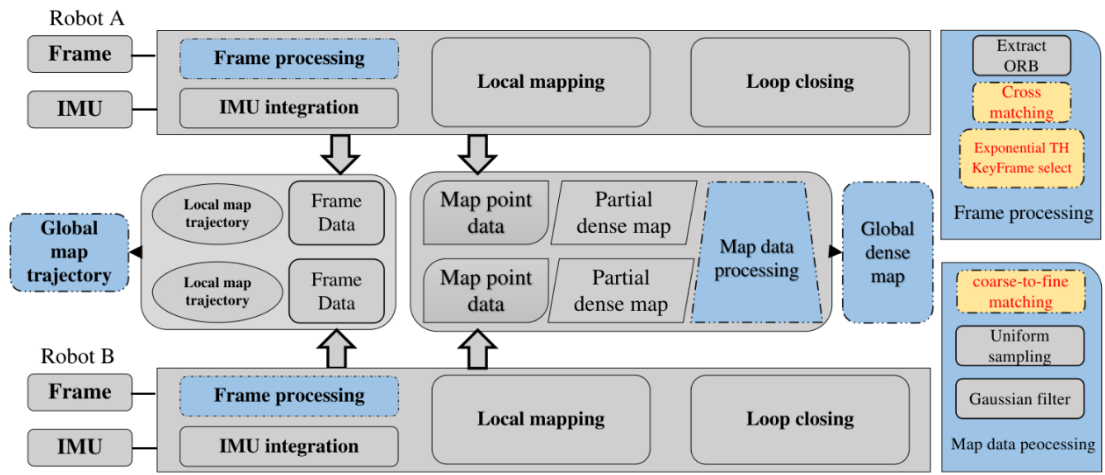

**Fig. 1** Overall system framework of multi-robot map building with improved feature matching and keyframe selection. The main improved modules are highlighted with a blue background, while the points of improvement are highlighted with a yellow background.

3.2. Feature matching

Expanding upon the original algorithm, this system introduces the cross-verification operation shown in Fig. 2 Instantiation provides examples of both match success and match failure.

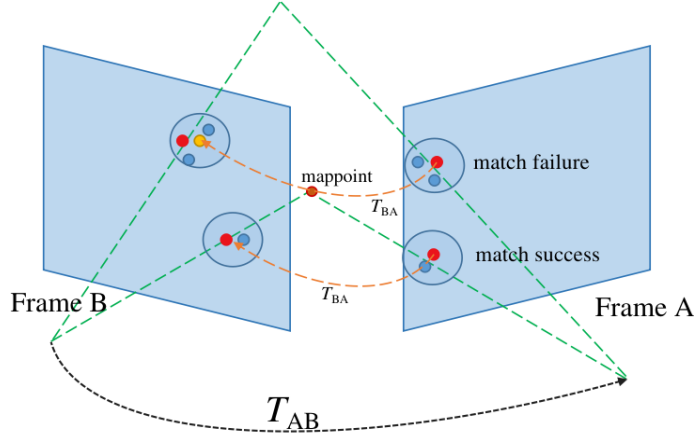

**Fig. 2** Schematic diagram of cross-validation-based feature matching

In the acceleration of feature matching based on grid registration information, we exemplify this through monocular initialization of matching between two frames. With the frames to be matched as Frame A and Frame B, their corresponding sets of feature points are denoted as $P_A = \{P_i^A \mid i = 1,2 \ldots, N_a\}$ and $P_B = \{P_i^B \mid i = 1,2 \ldots, N_b\}$, where $N_a$ ($N_b$) represents the number of feature points in $P_A$ ($P_B$). Here, $P_i^A$ ($P_i^B$) denotes the coordinates of the feature points in $P_A$ ($P_B$), where $P_i^A = (U_{ai}, V_{ai})$ and $P_i^B = (U_{bi}, V_{bi})$. The set of points $P_B$ in Frame B is projected onto the coordinate system of Frame A through a coordinate transformation $T_{AB}$, resulting in $P_{B_A}$:

$$P_{B_A} = T_{AB} P_B \qquad (1)$$

During the first stage of matching, the algorithm iterates through each point in $P_{B_A}$ and utilizes the Hamming distance between descriptors to determine the optimal matching point in the nearby grid region. By filtering out unmatched points, two initial matching point sets are obtained and denoted as $P'_A = \{P_i^{A'} \mid i = 1,2 \ldots, N'\}$ and $P'_B = \{P_i^{B'} \mid i = 1,2 \ldots, N'\}$. In this stage, points $P_i^{A'}$ in $P'_A$ and points $P_i^{B'}$ in $P'_B$ are one-to-one matched. The point set $P'_A$ from Frame A is subsequently projected onto the Frame B coordinate system through the transformation matrix $T_{BA}$ to obtain $P'_{A_B}$:

$$P'_{A_B} = T_{BA}P'_A \qquad (2)$$

During the second stage of matching, the algorithm iterates through each point in set $P'_{A_B}$ and finds the optimal matching point within the nearby grid region. The final sets of points that match each other are obtained as $P''_A = \{P_i^{A''} \mid i = 1,2 \dots, N''\}$ and $P''_B = \{P_i^{B''} \mid i = 1,2 \dots, N''\}$. In the experiments, a smaller matching radius can be adopted in this stage compared to the first stage of matching. This adjustment can filter out mismatching points more effectively, thus improving the matching results.

For the convenience of readers, the transformation matrix $T_{BA}$ is derived as follows:

$$T_{BA} = T_{BW} * T_{AW}^{-1} \qquad (3)$$

$$= \begin{bmatrix} R_{BW} & t_{BW} \\ 0^T & 1 \end{bmatrix} \begin{bmatrix} R_{AW}^T & -R_{AW}^T t_{AW} \\ 0^T & 1 \end{bmatrix}$$

$$= \begin{bmatrix} R_{BW} R_{AW}^T & -R_{BW} R_{AW}^T t_{AW} + t_{BW} \\ 0^T & 1 \end{bmatrix}$$

Here, $R_{Aw}(R_{Bw})$ represents the rotation matrix from the world coordinate system to Frame A (Frame B), and $t_{Aw}(t_{Bw})$ represents the translation vector from the world coordinate system to Frame A (Frame B). Considering that in the initial phase, the coordinates of the first frame are calibrated to the origin coordinates of the world coordinate system, $T_{AW} = T_{WA} = E$, where $E$ is the identity matrix. Thus, Eq (3) degenerates to:

$$T_{BA} = T_{AB}^{-1} = T_{WB}^{-1} = T_{BW} \qquad (4)$$

$$= \begin{bmatrix} R_{Bw} & t_{Bw} \\ 0^T & 1 \end{bmatrix}$$

In feature matching acceleration based on bag-of-words vectors, cross-validation is not suitable because the initial transformation matrix between two frames is not precomputed, and the feature point information in the key frame is transformed into a bag-of-words vector. Moreover, with the introduction of sliding windows and subpixel matching, the original feature matching algorithm achieves a remarkable balance between speed and accuracy in the acceleration of feature matching based on stripe search. There-fore, cross-validation is also not used in this stage because the introduction of cross-validation reduces the speed of binocular initialization in the stereo camera initialization stage.

3.3. Selection of the keyframe

Considering that the key frame selection strategy directly affects the positioning accuracy and mapping quality of the SLAM system, we assign different weights according to the influence of each factor. An exponential threshold function $T_H$ is constructed to quantify the selection of key frames:

$$T_H = \sum_{x \in X} W(e^x - 1) \tag{5}$$

We define $W$ as the weight matrix and $\Sigma$ as the covariance matrix encompassing all influencing factors. $X = [\alpha T_I, \beta R_A, \gamma T_R, \delta R_O]$, $T_I$, $R_A$, $T_R$, and $R_O$ denote the temporal cost, common view quality, translation cost, and rotational cost, respectively. $\alpha, \beta, \gamma$, and $\delta$ represent the weighting coefficients for each factor.

When the computed $T_H$ value of the current frame exceeds a predefined threshold $Q$, the current frame is designated a key frame, and it is advanced to the loop detection thread. At the same time, the current frame is updated as a reference key frame for subsequent processing. By constructing the exponential threshold function, the key frame selection factor can be nonlinearly and dynamic priority scheduling. This method quantifies the process of key frame selection and provides a new scheme for key frame selection, and its effect will also be reflected in the subsequent positioning accuracy and mapping effect.

3.4. Multi-maps matching and dense point cloud map construction

We make maximum use of the initial pose of each robot in the world coordinate system to construct a coarse-to-fine point cloud matching map. Then, the matched point cloud is processed via uniform sampling and Gaussian filtering to generate the final globally consistent dense point cloud map. To visually depict the delineated methodology, we showcase the flow of multi-map matching and dense point cloud map construction in Fig. 3.

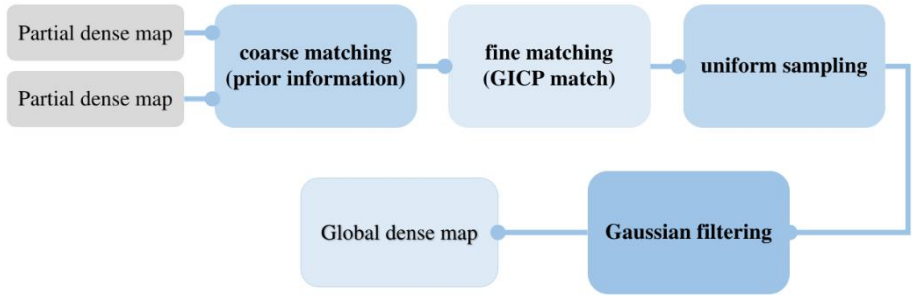

**Fig. 3** Multi-map matching and dense point cloud map building process

For simplicity, we take two robots, Robot A and Robot B, as examples for illustration. When each robot generates dense point cloud maps in their respective coordinate systems, the fusion of point cloud information becomes crucial. Therefore, we implement a coarse-to-fine matching approach.

During the coarse matching phase, we consider the initial positions of the robots and record the transformation matrices of each robot from the world coordinate system to their respective coordinate systems as $T_{AW}$ and $T_{BW}$. The point cloud collected by Robot A (B) is denoted as $P_A = \{P_i^A \mid i = 1,2 \ldots, N_a\}$ ($P_B = \{P_i^B \mid i = 1,2 \ldots, N_b\}$). Here, $N_a$ ($N_b$) represents the number of points in the point cloud $P_A$ ($P_B$). For each point, $P_i^A$ ($P_i^B$) represents the three-dimensional coordinates of the points in the point cloud under the Robot A (Robot B) coordinate system: $P_i^A = (X_{ai}, Y_{ai}, Z_{ai})$ ($P_i^B = (X_{bi}, Y_{bi}, Z_{bi})$). To measure them in the same coordinate system, we project the point clouds $P_A$ ($P_B$) onto the world coordinate system using the transformation matrices $T_{AW}$ and $T_{BW}$, which are denoted as $P_{A_W}$ ($P_{B_W}$):

$$P_{A_W} = T_{wA} P_A = T_{AW}^{-1} P_A \tag{6}$$

$$P_{B_W} = T_{WB} P_B = T_{BW}^{-1} P_B \tag{7}$$

Subsequently, by setting a matching radius to filter out unmatched points, we obtain two sets of mutually matched points denoted as $P'_{A_W} = \{P_i^{A'} \mid i = 1,2 \ldots, N'\}$ and $P'_{B_W} = \{P_i^{B'} \mid i = 1,2 \ldots, N'\}$. Here, $P_i^{A'}$ and $P_i^{B'}$ are neighboring points that are mutually matched. If $P'_{A_W}$ and $P'_{B_W}$ are not empty sets, a loop closure occurs in the trajectories of the two robots. It is worth noting that the value of $N'$ is less than that of $N_a$ and $N_b$ because only a portion of the points participate in the loop. If $N' = 0$, no points in the two sets of point clouds are involved in the loop. In such instances, it is impossible to proceed with the subsequent fine matching step, and $P_{A_W}$ and $P_{B_W}$ represent the final matching results of the point clouds.

During the fine matching phase, the generalized iterative closest point (GICP) [55] algorithm is employed, where the sets of points to be matched precisely are denoted as $P'_A$ and $P'_B$. Unlike the traditional ICP matching algorithm, the introduction of the covariance matrix helps eliminate the influence of certain undesired corresponding points during the solution process. In the GICP framework, it is assumed that the two sets of corresponding points, $P'_A$ and $P'_B$, follow a Gaussian distribution:

$$\begin{cases} P_i^{A'} \sim N(\hat{P}_i^{A'}, C_i^A) \\ P_i^{B'} \sim N(\hat{P}_i^{B'}, C_i^B) \end{cases} \quad (8)$$

$\hat{P}_i^{A'}$ and $\hat{P}_i^{B'}$ are the mean values in the Gaussian distribution, and $C_i^A$ and $C_i^B$ are the covariance matrices for each point. Leveraging the previously defined rigid transformation matrix $T_{BA}$:

$$T_{BA} = T_{BW} * T_{AW}^{-1} \quad (9)$$

The matching error $d_i$ for each pair of corresponding points is computed:

$$d_i = b_i - T_{BA} a_i \quad (10)$$

Given the independence of a and b, the distribution of d is also Gaussian:

$$d_i \sim N(0, C_i^B + T_{BA} C_i^A T_{BA}^T) \quad (11)$$

Therefore, the cost function for computing the transformation matrix is defined as follows:

$$T = \arg\min \sum d_i^T (C_i^B + T_{BA} C_i^A T_{BA}^T)^{-1} d_i \quad (12)$$

Through singular value decomposition (SVD), the precise transformation matrix between two robots can be determined.

Coarse-to-fine matching will generate a large amount of point cloud data, which poses a major challenge in terms of computing and storage resources. To solve this problem, uniform sampling [56] is used to reduce the amount of data, thus saving computational resources and mitigating the impact of noise and outliers. The algorithm's core step involves constructing a 3D voxel grid over the input point cloud data, and each voxel replaces all points within it with the point closest to the voxel center.

To further smooth the point cloud data and reduce the influence of noise points, we employed three-dimensional Gaussian filtering [57]. Each point $x_0$ in the point cloud map can be represented by the following formula:

$$f(x; x_0, \Sigma) = \frac{1}{(2\pi)^{\frac{d}{2}} |\Sigma|^{\frac{1}{2}}} \exp(x - x_0)^T \Sigma^{-1} (x - x_0) \quad (13)$$

where $x = \{x_i \mid i = 1, 2 \dots, 7\}$ denotes the coordinates of six points near point $x_0$ in three-dimensional space, $\Sigma$ is the covariance matrix, which represents the change relationship of point cloud data in each dimension, and $|\Sigma|$ denotes the determinant of the covariance matrix.

## 4. Experiments and analysis

To evaluate our system's performance, we use the TUM RGB-D [58] dataset as a benchmark. This dataset, sourced from the Technical University of Munich, is widely employed in computer vision, robotics, and autonomous navigation research. Its standardized motion trajectories make it ideal for assessing SLAM system localization accuracy. The experiments were run on an Intel(R) Core(TM) i9-12650H processor with 32 GB of RAM (Ubuntu 18.04). In the feature point matching experiments, we conducted comparisons between mainstream feature matching algorithms and the proposed cross-validation feature matching algorithm. In the keyframe selection experiments, we compared the localization accuracy of the proposed keyframe selection SLAM system based on exponential threshold function quantization against that of other advanced SLAM systems. In addition to the TUM RGB-D dataset, we utilized the EuRoC [59] and TUM_VI [60] datasets (Section 4.2) to validate the effectiveness and generalizability of the algorithm. Regarding multi-map matching, we presented the mapping results following coarse matching, fine matching, uniform sampling, and Gaussian filtering.

4.1. Feature point matching

When evaluating feature matching, metrics primarily focus on matching speed and accuracy. We carried out feature matching simulations from two sets of adjacent image frames randomly selected from the first and second halves of the fr1 xyz and fr3 teddy datasets, denoted as matches A, B, C, and D.

For each matched image set, 250, 500, 750 and 1000 feature points are extracted. Table 1 summarizes the comparison of the matching times of brute-force [61] matching, FLANN [62] matching, the original algorithm, and cross-validation matching. To reduce systematic errors, the matching process for each algorithm was repeated 100 times, and the results were averaged. Due to the lack of texture in match D, the ORB extraction method fails to extract 2000 feature points, rendering subsequent feature matching impossible. However, compared with those of the brute-force and FLANN algorithms, the matching speeds of the feature matching algorithm and cross-validation matching algorithm in ORB-SLAM3 are significantly improved when the number of extracted feature points exceeds 1000. This improvement stems from their utilization of feature matching acceleration based on grid registration information, which locates matching points near the same grid rather than searching through all images. Although the time consumption of the cross-validation matching algorithm is slightly greater than that of the original algorithm, it can still meet real-time performance requirements. More importantly, its matching accuracy is significantly improved, and the matching point distribution is more uniform.

**Table 1** Comparison of the runtimes of brute-force matching, FLANN matching, feature matching algorithms in ORB-SLAM3, and cross-validation matching.

| | match A | | | match B | | | match C | | | match D | | |
|---|---|---|---|---|---|---|---|---|---|---|---|---|
| Unit:(ms) | 250 | 1000 | 2000 | 250 | 1000 | 2000 | 250 | 1000 | 2000 | 250 | 1000 | 2000 |
| Brute-Force | 0.1742 | 1.6126 | 5.7742 | 0.1950 | 1.5463 | 5.6281 | 0.1845 | 1.6827 | 5.6302 | 0.1945 | 1.5826 | - |
| FLANN | 0.6328 | 2.9502 | 8.4999 | 0.6279 | 2.9194 | 7.9627 | 0.4964 | 3.3381 | 9.0401 | 0.5236 | 2.9554 | - |
| ORB-SALM3 | 0.2581 | 1.0461 | 2.2981 | 0.2652 | 0.7832 | 1.6840 | 0.4387 | 1.5835 | 3.0562 | 0.3023 | 1.1764 | - |
| Cross-validation | 0.3036 | 1.2405 | 2.5741 | 0.2771 | 1.0338 | 2.2671 | 0.4037 | 1.7223 | 3.5821 | 0.4219 | 1.482 | - |

In extensive feature matching experiments, we compare two typical feature matching performances of the ORB-SLAM3 algorithm (Fig. 4(a, c)) and the cross-validation matching algorithm (Fig. 4(b, d)). For the improved cross-validation matching algorithm, the mismatch points are filtered after cross-validation, making the distribution of feature points more uniform, as shown in Fig. 4(b). However, compared with the image in Fig. 4(c), the feature matching effect with the cross-validation matching algorithm is not significantly enhanced in Fig. 4(d), which can be attributed to the fact that the original algorithm generates a uniform distribution of feature points.

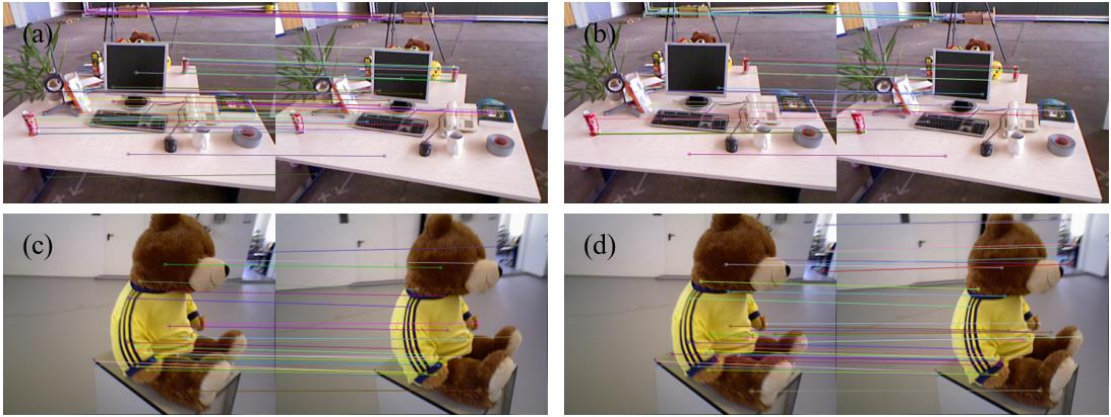

**Fig. 4** Feature matching results for adjacent image frames: (**a, c**) original algorithm; (**b, d**) cross-validation matching algorithm.

4.2. Keyframe Selection

The effectiveness of localization and mapping is influenced by various factors, including the quality of feature matching, keyframe selection strategies, and the quality of local and global bundle adjustment optimization. In our experiments, we observed that improving cross-validation feature matching had a significant impact on mapping effectiveness but had a minor impact on localization effectiveness. Additionally, the selection of keyframes notably influenced both localization accuracy and mapping effectiveness. Therefore, in subsequent experiments, we reverted to using the original feature matching scheme, focusing solely on verifying the localization effectiveness through the improvement of keyframe selection strategies.

To provide a comprehensive assessment of the localization accuracy of ORB-SLAM2, the original ORB-SLAM3 algorithm, and the enhanced ORB-SLAM3 algorithm incorporating an exponential threshold function for keyframe determination, we conducted multiple experiments using the TUM RGB-D, EuRoC MAV, and TUM_VI datasets. Utilizing the EVO [58] evaluation tool, we employ the absolute trajectory error (ATE) [58] as the evaluation metric to compare the results of the algorithm's execution with the ground truth provided by the dataset.

4.2.1. Visual comparison of algorithm trajectories with reference trajectories

To validate the improved algorithm, experiments were conducted on all sequences of the TUM RGB-D dataset captured in RGB-D mode. We selected representative sequences, fr1 xyz and fr3 teddy, for demonstration. The fr1 xyz dataset comprises an office desktop and surrounding scenes captured by a Kinect camera. The fr3 teddy dataset consists of teddy bears and their surrounding scenes captured by an Xtion camera at different heights in an indoor environment.

Fig. (5) and (6) show the test results on the fr1 xyz and fr3 teddy datasets, respectively. Fig. 5(a-c) and 6(a-c) depict the three-dimensional trajectory comparisons between the algorithms' running trajectories and the ground truth trajectories. Fig. 5(d-f) and 6(d-f) show the variations in the trajectory errors over time. For the ORB-SLAM2 algorithm, in the fr1 xyz dataset, there is a significant turning amplitude of the camera at approximately 3-3.5 seconds. During this period, the algorithm fails to insert and process key frames in a timely manner, causing the motion model to lose track. Consequently, there is considerable error between the tracked trajectory and the ground truth trajectory. However, correct trajectory tracking is achieved by tracking the reference keyframe and BA pose graph optimization, as shown in Fig. 5(a, d). In the fr3 dataset, similar results are observed, and a significant trajectory deviation occurs at approximately 50-63 seconds, as depicted in Fig. 6(a, d). For the ORB-SLAM3 algorithm, compared with the running results of ORB-SLAM3 illustrated in Fig. 5(b, e), the running trajectory of the improved algorithm depicted in Fig. 5(c, f) is closer to the true trajectory, and the maximum value of APE is reduced by 0.006 m. The running results on the fr3 teddy dataset more obviously reflect the advantages of the improved algorithm. The ORB-SLAM3 algorithm exhibited in Fig. 6(b, e) has insufficient quality and quantity of key frame selection in high-frequency signal scenes, such as high-intensity turning within 62 to 75 s, and the motion trajectory has a large deviation from the real trajectory. The improved algorithm, shown in Fig. 6(c, f), sets reasonable weights for the translation factor and rotation factor, so the keyframes can be added to the loop thread in time in the scene of high-intensity turning, which significantly diminishes the positioning error of the SLAM system, ensuring the quality of the subsequent map.

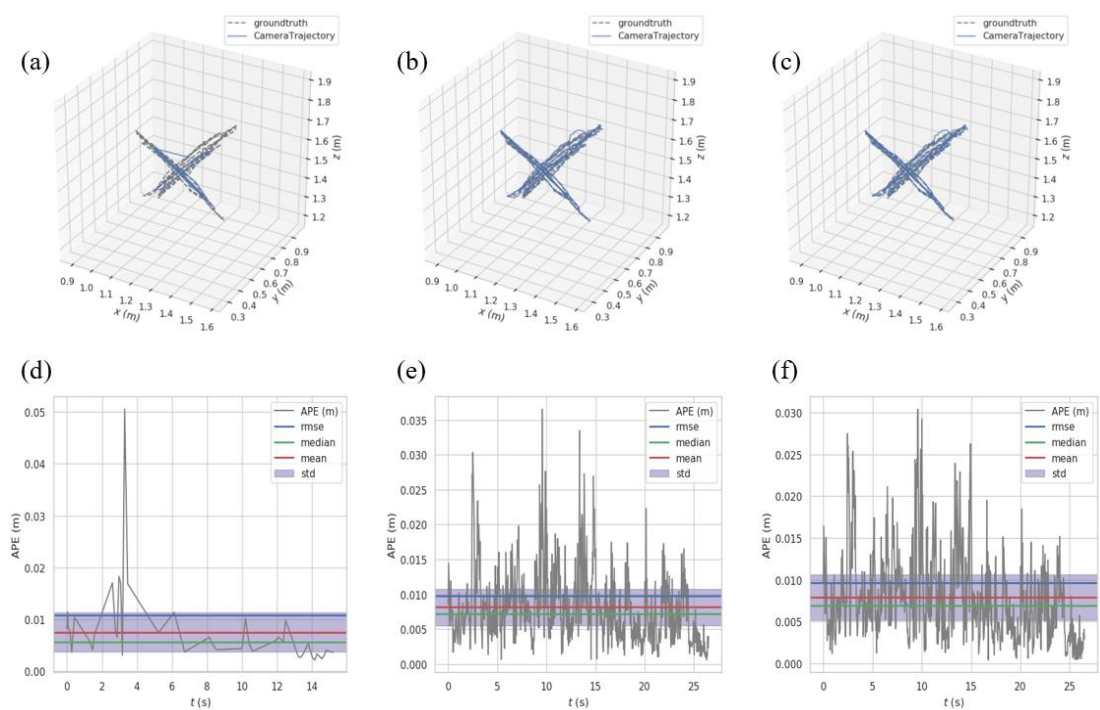

**Fig. 5** The motion trajectory and absolute trajectory error obtained from the fr1 xyz dataset: (a, d) ORB-SLAM2 algorithm; (b, e) ORB-SLAM3 algorithm; (c, f) improved algorithm.

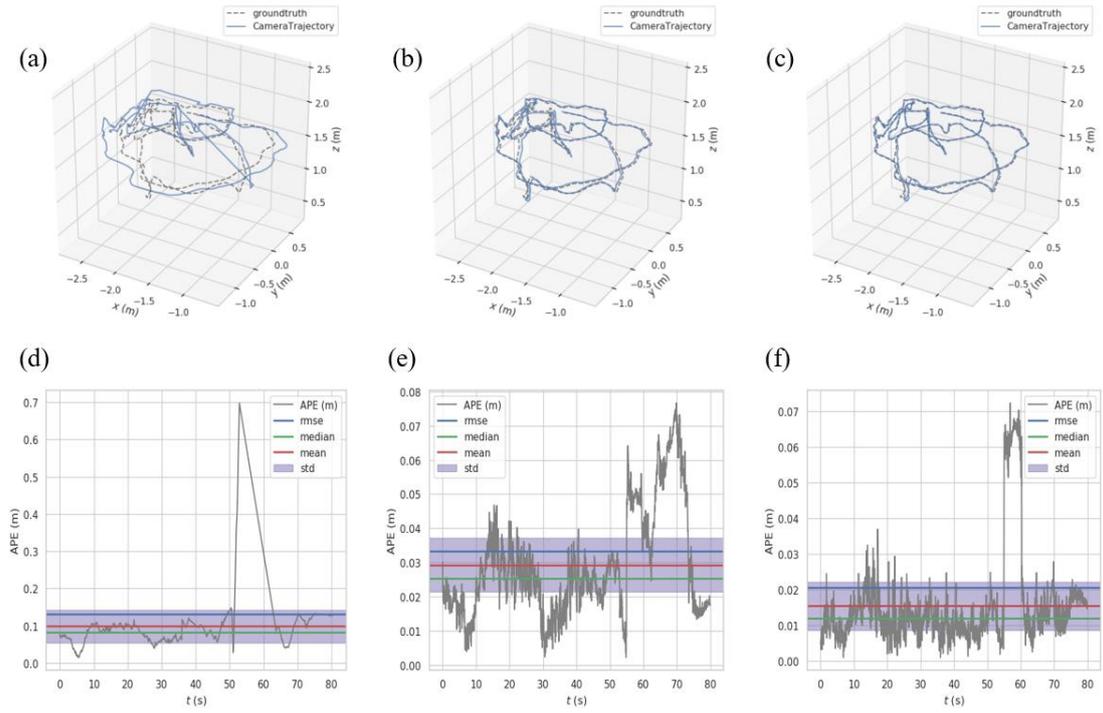

**Fig. 6** The motion trajectory and absolute trajectory error obtained from the fr3 teddy dataset: (a, d) ORB-SLAM2 algorithm; (b, e) ORB-SLAM3 algorithm; (c, f) improved algorithm.

4.2.2. Quantitative comparison of algorithm trajectories with reference trajectories

We selected the fr1 xyz, fr3 teddy, and fr1 desk2 sequences from the TUM RGB-D dataset and compared the ATEs of ORB-SLAM2, ORB-SLAM3, and the improved algorithms. To mitigate the impact of systematic errors, each sequence was iterated 10 times, excluding the maximum and minimum values and averaging the remaining results. The evaluation metrics included the maximum and median errors, root mean square error (RMSE), and standard deviation (STD). The results are presented in Table 2. Experiments on the fr1 xyz and fr3 teddy datasets show that the improved algorithm achieves the best performance in all evaluation indicators and is significantly better than ORB-SLAM2. Especially on the fr3 teddy dataset, in comparison to the original ORB-SLAM3, the improved algorithm exhibits superior positioning accuracy across all evaluation indices. When tested on the fr1 desk2 dataset, the improved algorithm and the original ORB-SLAM3 algorithm have their own advantages and disadvantages according to different evaluation indicators. In general, on this test dataset, compared with the original ORB-SLAM3 algorithm, the maximum absolute trajectory error of the improved algorithm is reduced by 10.25%, the median trajectory error is reduced by 9.01%, the RMSE is reduced by 12.90%, and the STD is improved by 4.86%, significantly improving the positioning accuracy.

**Table 2** Comparison of the ATEs of ORB-SLAM2, ORB-SLAM3, and improved algorithms with the exponential threshold function. The best ATE results are shown in bold.

|          | fr1 xyz   |           |          | fr3 teddy |           |          | fr1 desk2 |           |          |
|----------|-----------|-----------|----------|-----------|-----------|----------|-----------|-----------|----------|
| Unit:(m) | ORB_SLAM2 | ORB_SLAM3 | Ours     | ORB_SLAM2 | ORB_SLAM3 | Ours     | ORB_SLAM2 | ORB_SLAM3 | Ours     |
| max      | 0.067805  | 0.036584  | **0.030568** | 0.590751 | 0.077683 | **0.072839** | 2.436221 | 0.064827 | **0.059601** |
| med      | 0.007890  | 0.007112  | **0.006850** | 0.065815 | 0.025283 | **0.012110** | 0.549178 | **0.016545** | 0.021305 |
| RMSE     | 0.014579  | 0.009715  | **0.009677** | 0.112216 | 0.033576 | **0.020592** | 0.852212 | **0.020864** | 0.024506 |
| STD      | 0.010298  | 0.005293  | **0.005241** | 0.074586 | 0.015718 | **0.013549** | 0.525632 | 0.010437 | **0.010416** |

Fig. 7 illustrates the cumulative ATE of the fr3 teddy sequence from the TUM RGB-D dataset over multiple runs with a runtime of 80.8 s. The cumulative results using the RMSE and STD as evaluation metrics are shown in Fig. 7(a) and Fig. 7(b). These two cumulative ATE plots demonstrate that with increasing running time, the positioning accuracy of the proposed algorithm is obviously superior to that of other algorithms.

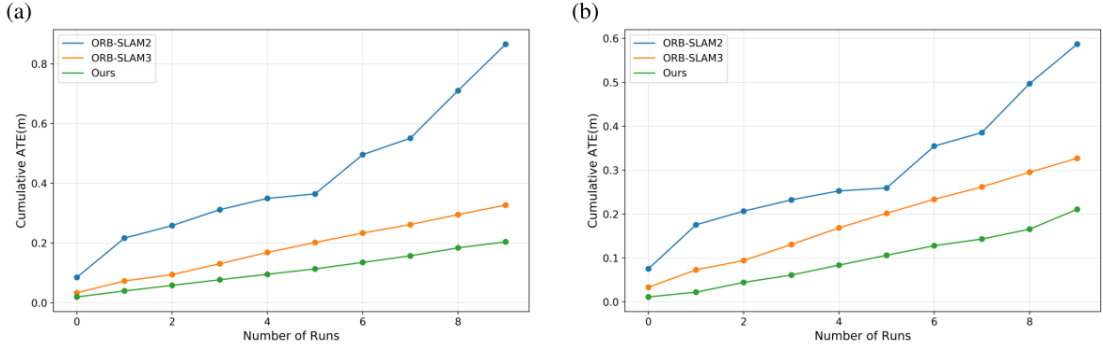

**Fig. 7** Cumulative ATE results of the fr3 teddy datasets in the rgb-d mode: (a) RMSE as an evaluation metric; (b) STD as an evaluation metric.

To evaluate the generalization capability of the improved algorithm, we conducted additional tests on both the EuRoC MAV [59] dataset and the TUM_VI [60] dataset. Fig. 8 displays the test results obtained from the EuRoC MAV dataset, while Fig. 9 depicts the outcomes from the TUM_VI dataset. The improved algorithm yields lower RMSE and standard deviation values than does the ORB-SLAM3 algorithm. This trend is particularly pronounced in the mono-inertial mode, as observed in sequences such as MH4 in the EuRoC MAV dataset and ROOM1 and ROOM4 in the TUM_VI dataset. In these datasets, the improved algorithm effectively mitigates long-term trajectory deviations caused by the absence of appropriate keyframe tracking, thereby reducing the likelihood of significant trajectory divergence. These results further emphasize the advantages of the algorithm in terms of the positioning accuracy and generalizability of different datasets.

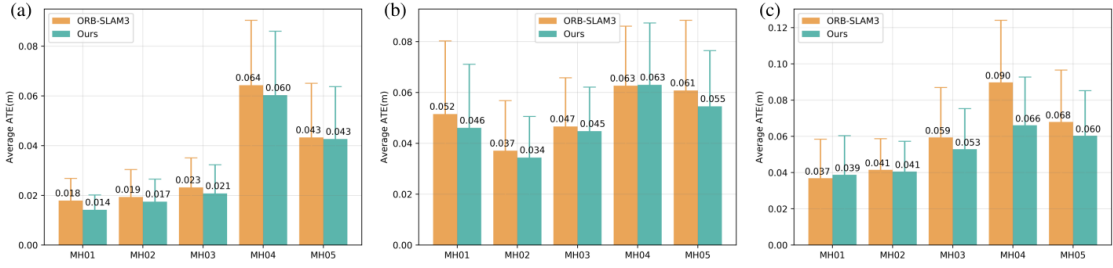

**Fig. 8** Average RMSE and STD results for the MH sequence in the EuRoC dataset: (a) Stereo model; (b) Stereo-inertial mode; (c) Mono-inertial mode;

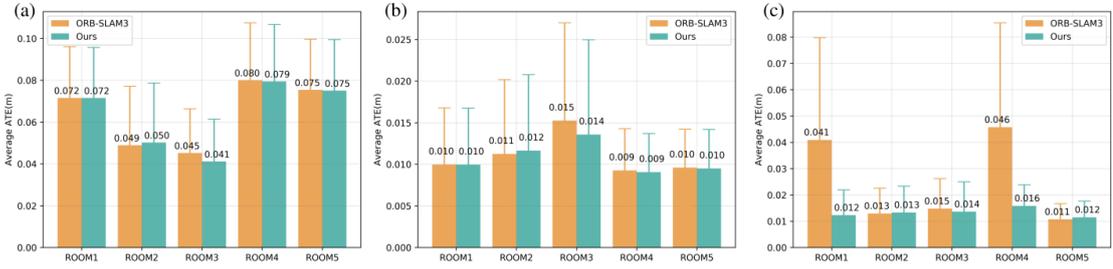

**Fig. 9** Average RMSE and STD results for the ROOM sequence in the TUM_VI dataset: (a) Stereo model; (b) Stereo-inertial mode; (c) Mono-inertial mode;

4.3. Multimap matching

The ORB-SLAM3 algorithm is a classical representative feature point method, but it has one limitation: it can only generate sparse point cloud images. Based on the original algorithm, we add a dense mapping thread and carry out multi-map matching through coarse-to-fine matching. Thus, a complete multi-robot SLAM system is constructed. Fig. 10 depicts the validation results of this proposed system utilizing the fr1 xyz dataset, retaining the intermediate steps of algorithm execution. Fig. 10(a, b) shows the dense point cloud maps constructed by robots A and B, respectively. Fig. 10(c) depicts the coarse-to-fine matching process, where the green segment represents the contribution of robot A's dense point cloud image to the global map, and the red segment represents the contribution of robot B's dense point cloud image. Fig. 10(d) shows the global point cloud image after fusion. In addition, we also observed that the point cloud of the fused map contained an excessive number of points. To address this, we introduce a subsequent uniform sampling module, as depicted in Fig. 10(e). The module significantly reduced the data density while preserving the key characteristics of the point cloud data. Comparing Fig. 10(d) and 10(e) before and after the uniform sampling operation, we observed that the mapping did not deteriorate. Fig. 10(f) illustrates the result after Gaussian filtering. The filtering step effectively mitigates the influence of noise points on the point cloud map, resulting in smoother point cloud data. Notably, components such as the desktop and ground in the point cloud map exhibit enhanced smoothness and better approximation to the real scene, improving the mapping quality.

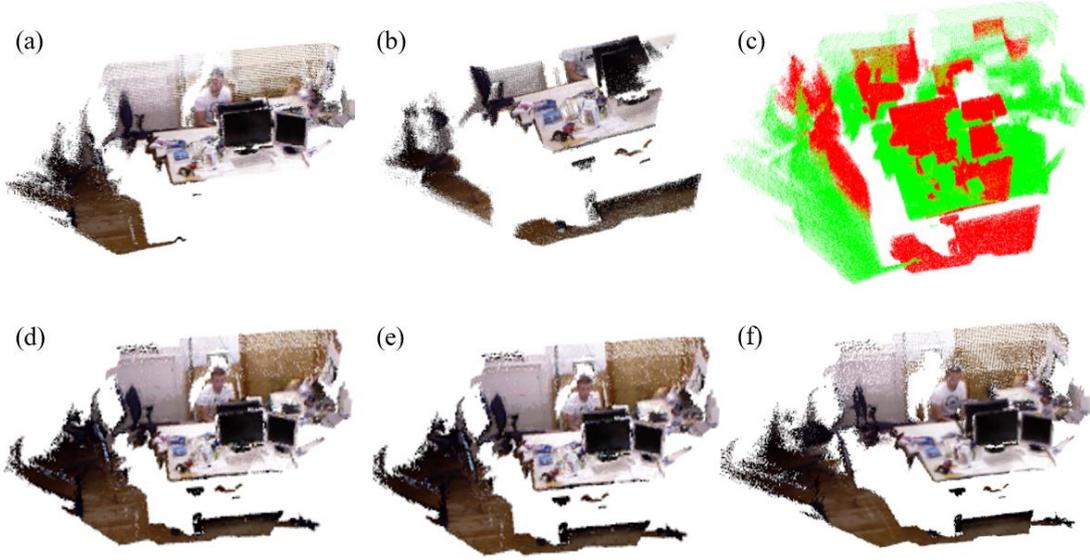

**Fig. 10** Process of multi-robot construction of a dense point cloud map on the fr1 xyz dataset

Table 3 provides a summary of the comparisons before and after uniform sampling. These experiments revealed a decrease in the number of points in the point cloud by an average of 32.89%. In the fr1 xyz and fr1 desk2 datasets, which are characterized by relatively small environments, a substantial portion of the point cloud data acquired by both robots correspond to the same scene. Consequently, the uniform sampling process effectively filters out a significant portion of the overlapping point clouds. Conversely, the fr3 teddy scene is more spacious, resulting in fewer overlapping regions scanned by both robots and consequently fewer adjacent points in the point cloud. Thus, the percentage decrease in points after uniform sampling is notably smaller than that in the aforementioned datasets.

**Table 3** The number of points in the point cloud before and after uniform sampling and the decrease ratio of the number of point clouds after uniform sampling

|  | **fr1 xyz** | **fr1 desk2** | **fr3 teddy** | **Ours dataset** |
|---|---|---|---|---|
| dataset length(s) | 30.093 | 24.861 | 80.789 | 33.984 |
| before sampling | 206416 | 907627 | 2942614 | 258352 |
| after sampling | 131537 | 597264 | 2355203 | 154685 |
| reduction (%) | 37.28 | 34.19 | 19.96 | 40.13 |

## 5. Conclusions

In this work, we introduce a novel feature matching and keyframe selection approach that incorporates cross-validation into the feature matching process and constructs an exponential threshold function for keyframe determination. Based on these innovations, we compare the improved ORB-SLAM3 with mainstream algorithms, achieving notable results such as a significant reduction in mismatch rates and a 12.90% enhancement in the overall localization accuracy of the SLAM system. In addition, a centralized multi-robot collaboration scheme is used to collect the point cloud map built by each robot. A coarse-to-fine multi-map matching scheme is designed to achieve accurate map matching, followed by postprocessing steps such as uniform sampling and Gaussian filtering. Experimentally, our system can complete the multi-robot CSLAM task well and achieve high-precision map reconstruction.

In the future, we plan to explore distributed and hybrid CSLAM schemes. In addition, since there are a large number of dynamic objects in an actual scene, which will affect the positioning and mapping effect of the system, we also consider using a combination of geometric structure and deep learning to detect and filter dynamic features.

**Acknowledgments** This work was supported by the National Natural Science Foundation of China (No. 11904056). Guangzhou Basic and Applied Basic Research Project (No. 202102020053).

**Author Contributions** Conceptualization, X.-b.G.; data curation and software, A.-H., X.-m.W. and X.-b.G.; funding acquisition, X.-b.G.; investigation, A.-H., X.-m.W., Z.- S., L.-b.L., and X.-b.G.; resources, L.-b.L.; supervision, X.-b.G. and L.-b.L.; writing—original draft, A.-H., X.-m.W., and Z.- S. ; writing—review and editing, W.-h.Qiu., L.-b.L., and X.-b.G.; All authors have read and agreed to the published version of the manuscript.

**Data Availability Statement** The TUM RGB-D dataset is obtained from https://cvg.cit.tum.de/data/

.datasets/rgbd-dataset/download.The EuRoC MAV dataset is obtained from https://projects.asl.ethz.ch/datasets/doku.php?id=kmavvisualinertialdatasets. The TUM_VI datasets is obtained from https://cvg.cit.tum.de/data/datasets/visual-inertial-dataset. Our real-world test is available upon request from the corresponding author.

**Code Availability** The code generated during the current study will be refined and then be available on GitHub.